\documentclass[conference]{IEEEtran}
\IEEEoverridecommandlockouts
\usepackage{cite}
\usepackage{amsmath,amssymb,amsfonts,amsthm}
\usepackage{algorithmic}
\usepackage{graphicx}
\usepackage{textcomp}
\usepackage{natbib}
\usepackage{xcolor}
\usepackage{mathtools}
\usepackage{hyperref}
\usepackage{subcaption}
\def\BibTeX{{\rm B\kern-.05em{\sc i\kern-.025em b}\kern-.08em
    T\kern-.1667em\lower.7ex\hbox{E}\kern-.125emX}}

\renewcommand{\cite}{\citep}
\newcommand{\dd}{\mathrm{d}}
\DeclareMathOperator{\diag}{diag}
\newtheoremstyle{regularTheorem}
  {\topsep}    % space above
  {\topsep}    % space bellow
  {\itshape}   % body font†
  {}           % no indent
  {\bfseries}  % head font
  { }          % char after heading 1
  { }          % char after heading 2
  {}           % misc
\theoremstyle{regularTheorem}
\newtheorem{definition}{Definition}

\begin{document}

\title{A Truly Sparse and General Implementation of Gradient-Based Synaptic Plasticity$^*$\\
\thanks{$^*$Sponsored by the Federal Ministry of Education and Research BMBF Neurosys and Neurotec II (03ZU1106CB, 16ME0398K, and 16ME0399).}
}

\author{\IEEEauthorblockN{Jamie Lohoff\(^\dagger\)}
\IEEEcompsocitemizethanks{\(\dagger\)These authors contributed equally.}
\IEEEauthorblockA{\textit{Peter Gr\"unberg Institute 15} \\
\textit{FZ J\"ulich, RWTH Aachen} \\
Aachen, Germany \\
ja.lohoff@fz-juelich.de}
\and
\IEEEauthorblockN{Anil Kaya\(^\dagger\)}
\IEEEauthorblockA{\textit{Peter Gr\"unberg Institute 15} \\
\textit{FZ J\"ulich, RWTH Aachen}\\
Aachen, Germany \\
a.kaya@fz-juelich.de}
\and
\IEEEauthorblockN{Florian Assmuth}
\IEEEauthorblockA{\textit{Peter Gr\"unberg Institute 15} \\
\textit{FZ J\"ulich, RWTH Aachen}\\
Aachen, Germany \\
f.assmuth@fz-juelich.de}
\and
\IEEEauthorblockN{Emre Neftci}
\IEEEauthorblockA{\textit{Peter Gr\"unberg Institute 15} \\
\textit{FZ J\"ulich, RWTH Aachen} \\
Aachen, Germany \\
e.neftci@fz-juelich.de}}

\maketitle

\begin{abstract}
Online synaptic plasticity rules derived from gradient descent achieve high accuracy on a wide range of practical tasks. 
However, their software implementation often requires tediously hand-derived gradients or using gradient backpropagation which sacrifices the online capability of the rules.
In this work, we present a custom automatic differentiation (AD) pipeline for sparse and online implementation of gradient-based synaptic plasticity rules that generalizes to arbitrary neuron models. 
Our work combines the programming ease of backpropagation-type methods for forward AD while being memory-efficient.
To achieve this, we exploit the advantageous compute and memory scaling of online synaptic plasticity by providing an inherently sparse implementation of AD where expensive tensor contractions are replaced with simple element-wise multiplications if the tensors are diagonal.
Gradient-based synaptic plasticity rules such as eligibility propagation (e-prop) have exactly this property and thus profit immensely from this feature.
We demonstrate the alignment of our gradients with respect to gradient backpropagation on an synthetic task where e-prop gradients are exact, as well as audio speech classification benchmarks. 
We demonstrate how memory utilization scales with network size without dependence on the sequence length, as expected from forward AD methods.
\end{abstract}


\section{Introduction}
Recent advances in deep learning continually push the boundary of tackling and solving new challenges previously thought to be unsolvable.
However, deep learning achieves its performance through gradient backpropagation which is not biologically plausible.
Backpropagation violates two fundamental properties of physical computing, which are spatial locality and online learning.
In the brain, synaptic plasticity plays an important role in achieving mid- to long-term adaptation of behavior.
A common ground between biological heuristics and optimization algorithms in deep learning is provided by synaptic plasticity rules derived from gradient descent. 
However, in many cases the implementation of these gradient-based plasticity rules sacrifices generality by being meticulously handcrafted for a certain problem and neuron model.
Machine learning software frameworks are generally designed for efficient execution of backpropagation on various general purpose accelerators.
Synaptic plasticity rules derived from gradient descent can leverage these frameworks and backpropagation\footnote{Often with \textit{stop\_gradient} operations where necessary to take into account the approximations used in the plasticity rule} at the cost of sacrificing online operation and requiring memory that grows with sequence length.

However, gradient-based plasticity rules operate on the principle of forward-mode Automatic Differentiation (AD) \cite{Zenke_Neftci21_brailear} with approximations that constrain the rule to be local \cite{Murray19_locaonli}. 
In the case of recurrent neural networks that are unrolled along the time axis, forward-mode AD is also called Real-Time Recurrent Learning (RTRL) \cite{Williams_Zipser89_learalgo}.
With RTRL, the information necessary for computing gradients is propagated forward in time, allowing for online learning.
However, na\"ive implementations of forward-mode AD have high memory and time complexity compared to backpropagation (\textit{reverse-mode AD}) \cite{Williams_Zipser95_gradlear}.
% Such na\"ive implementations ignore the local approximations in the synaptic plasticity rules, leading to a significant overhead.
Such na\"ive implementations calculate the exact gradients, and miss the improvements gained with local approximations.

By providing a novel extension of the popular machine learning framework JAX \cite{jax2018github} this work aims to harnesses the advantageous properties of approximate gradient-based synaptic plasticity rules.
%As discussed in previous work, gradient-based synaptic plasticity rules have a huge potential for optimization because these
We employ the recently proposed AD library, Graphax \cite{Lohoff_Neftci24_optiauto} to showcase the efficient implementation of gradient-based plasticity rules and exploiting their inherently sparse formulation to arrive at the computational and memory complexities as expected from theory (see \autoref{tab:ComputeMemoryComplexity}).
Our contributions are summarized as follows:
\begin{itemize}
\item We provide the first AD package for spiking neural networks that leverages the inherent sparsity of gradient-based synaptic plasticity rules,
%% Following is slippery: your use cases are in a computer and uses batches, so it is not 100% online
%\item Truly online,
\item Our AD package can automatically differentiate through arbitrary neuron dynamics and determine the resulting eligibility traces and their sparsity structure,
\item We demonstrate how custom sparse AD operates faster and more memory efficient than current state-of-the-art BPTT methods.
\end{itemize}
The source code of the project is available under \href{https://github.com/jamielohoff/synaptax}{https://github.com/jamielohoff/synaptax}.
\begin{table}[htbp]
\caption{Upper bounds on computational and memory complexity for various gradient-based synaptic plasticity algorithms.}
    \scriptsize
    \begin{center}
        \begin{tabular}{|l|c|c|c|c|c|}
            \hline
             & RTRL & BPTT & e-prop (BPTT) & \textbf{e-prop} & \textbf{Ours}\\
            \hline
            Compute & \( \mathcal{O}(n^4T) \) & \( \mathcal{O}(n^2T) \) & \( \mathcal{O}(n^2T) \)& \( \boldsymbol{\mathcal{O}}(\boldsymbol{n^2}T) \) & \( \boldsymbol{\mathcal{O}}(\boldsymbol{n^2T}) \) \\
            \hline
            Memory & \( \mathcal{O}(n^3) \) & \( \mathcal{O}(nT) \) & \( \mathcal{O}(nT) \) & \( \boldsymbol{\mathcal{O}}(\boldsymbol{n}) \) & \( \boldsymbol{\mathcal{O}}(\boldsymbol{n}) \) \\
            \hline
            Automatic & Yes\footnotemark & Yes & Yes & \textbf{No} & \textbf{Yes} \\
            \hline
        \end{tabular}
    \label{tab:ComputeMemoryComplexity}
    \end{center}
\end{table}
\pagebreak
\section{Prior Work}
\footnotetext{e.g. in JAX using \texttt{jacfwd}}
Using various forms of surrogate functions, a wide range of methods have been developed for training spiking neural networks that approximate gradient backpropagation \cite{Bohte_etal00_spikback,Shrestha_Orchard18_slayspik,Neftci_etal19_surrgrad}.
In shallow, feed-forward networks, these gradient-based rules can be written in an online form \cite{Zenke_Ganguli18_supesupe,Kaiser_etal20_synaplas,Bellec_etal20_soluto} that can be interpreted as a three-factor rule \cite{Gerstner_etal18_eligtrac}.
In deep networks, the spatial credit assignment problem can be addressed by backpropagating in space  and forward propagating in time \cite{Kaiser_etal20_synaplas,Bohnstingl_etal20_onlispat}.
Such synaptic plasticity rules generally have to be hand-crafted, especially for the forward propagation.
While some methods have shown that temporal gradients can be ignored \cite{Yin_etal23_accuonli,Meng_etal23_towamemo}, these either use other approximate methods for computing traces or test on inherently static datasets where temporal integration is not necessary. 
%% The following paragraph may not be necessary. Depends if these can be accelerated with Graphax
Other approaches that do not use gradient backpropagation are \textit{event-prop}, which uses the adjoint method to compute gradients exactly when the number of events is fixed \cite{Wunderlich_Pehle21_evenback}; parameter updates based on spike times \cite{Goltz_etal21_fastener}; and prospective coding which leverages the advanced response of the neuron with respect to their inputs \cite{Brandt_etal24_prosretr}. 

To study gradient-based synaptic plasticity rules, several software frameworks for SNN training have been developed.
These are typically built on top of common machine learning libraries like PyTorch, TensorFlow or JAX.
Particularly PyTorch is very popular for implementation of such simulators with \textit{snnTorch} and \textit{SpikingJelly} being the most prominent \cite{Eshraghian_etal23_traispik,Fang_etal23_spikopen}.
Additionally, \textit{Norse} implements a spike-time based approach to gradient computation \cite{Pehle_Pedersen21_nors}.
These simulators enable simulation and training of SNNs with almost arbitrary network topology and neuron types on modern general purpose accelerators like GPUs.
Another strain of frameworks puts more emphasis on the training of SNNs dedicated to specific neuromorphic hardware backends, thereby (sometimes artificially) limiting the set of operations.
Examples include \textit{Lava} and \textit{SINABS} \cite{sinabs, lava2021github}.
Other works have focused on accelerating gradient computations since they pose a significant bottleneck to SNN training where sequence lengths are generally in the hundreds.
They leverage implicit differentiation and custom CUDA kernels for simple LIF models \cite{Bauer_etal22_exodstab}.

However, these methods require meticulously handcrafted kernels for every neuron type which makes them often unfeasible in practice.
Recent works instead implemented SNN simulators in JAX, a general-purpose numerical computing library with \textit{numpy}-like API.
JAX-based simulators like \textit{SNNAX}, \textit{Slax} and \textit{Spyx} take advantage of Just-In-Time (JIT) compilation to accelerate training, thereby achieving similar performance as methods with custom kernels \cite{Lohoff_etal24_snnaspik, kade_heckel_2024_10635178, summe2024slaxcomposablejaxlibrary}.
One disadvantage of these works is the lack of interoperability with existing neuromorphic hardware, although initiatives like the Neuromorphic Intermediate Representation (NIR) aim to alleviate this issue \cite{pedersen2024}.

Given the inherent inefficiency of gradient backpropagation in SNNs, several research efforts have investigated gradient computations with sparse activations \cite{PerezNieves_Goodman21_sparspik}. 
However, taking advantage of this sparsity requires specialized accelerators and optimizing custom kernels \cite{Finkbeiner_etal24_harnmany}. 
% Furthermore, many of the investigated approaches still utilize backpropagation and thus scale with sequence length and are inherently offline.
Nearly all SNN simulators implement gradient-based plasticity rules using backpropagation, either because the underlying software framework only has limited support for forward propagation (e.g. PyTorch) or because it ignores the sparsity patterns induced by the approximations.
Furthermore, this dependency on backpropagation forces these implementations to be inherently offline.
Our work tackles both of these issues by using the recently published \textit{Graphax} AD package to exploit the inherent sparsity patterns of popular gradient-based synaptic plasticity rules.
At the same time our is automatically computes traces and their sparsity patterns for arbitrary neuron types using a AD method \textit{vertex elimination}.
This leads to a significant reduction in runtime and memory cost as shown in \autoref{tab:ComputeMemoryComplexity}.
\section{Gradient-based Synaptic Plasticity Rules and Automatic Differentiation}
\label{sec:Eprop}
In machine learning, parameters are typically updated by computing gradients with respect to some cost function.
AD computes gradients by first decomposing the computational graph of the desired model into its elemental operations whose derivatives are known and implemented into the respective framework.
The gradients are then accumulated using the chain rule.
Fundamentally, this involves only summation and product operations acting on the derivatives of the elemental functions.
The result is the Jacobian or gradient of the respective model with respect to its parameters.
Note that AD computes the Jacobian up to machine precision since there are no approximations made on the known partial derivatives or their accumulation at any stage.

Forward-mode and reverse-mode are the two most popular AD methods with reverse-mode being synonymous with backpropagation.
Their main difference lies in the way the two methods traverse the computational graph to accumulate the gradient.
Forward-mode traverses the computational graph in the order of execution of the elemental operations.
Reverse-mode traverses the computational graph in the opposite direction, thereby propagating the error signal from the cost function through the network.
The direction of traversal has significant impact on the compute and memory requirements of both algorithms.
For an arbitrary function \( f: \mathbb{R}^n \to \mathbb{R}^m \) forward-mode operates better with larger outputs than inputs (\(m\gg n\)) while reverse-mode operates fewer computations with a larger inputs than outputs (\(n\gg m\)). 
This is especially important for  ``'funnel-like' computational graphs such as deep neural networks with many inputs and only a single output. 
Although forward mode does not require storing intermediate variables for the backward pass, it incurs a high memory cost in most practical problems and a significant computational overhead when \( m \gg n \) which limits its practical use.
Thus, backpropagation has established itself as the pillar of modern deep network training.%, as its advantageous compute scaling outweigh the memory overhead.
\subsection{The Role of Time and Recursion Relations}
Allowing the network to evolve in time, as for the case in Recurrent Neural Networks (RNNs), adds a new dimension to the development of gradient-based learning algorithms.
Typically modern RNNs are modeled by discretizing time and binning data into time frames.
In this case the general RNN recursion relation reads
\begin{equation}
    \boldsymbol{h}_{t+1} = \boldsymbol{f}(\boldsymbol{h}_t, \boldsymbol{x}_t, \theta),
\end{equation}
where \(\boldsymbol{h}_t \in \mathbb{R}^n\) denotes the hidden states, \(\boldsymbol{x}_t \in \mathbb{R}^k\) denotes external inputs, and \(\theta \in \mathbb{R}^p\) are the models parameters.
These hidden states can be unrolled and treated similarly as a deep network.
To compute the gradient, we run forward-mode or reverse-mode AD over the unrolled computational graph and arrive at Real-Time Recurrent Learning (RTRL) or Backpropagation Through Time (BPTT), respectively.
The general mathematical formulation of RNN training with a cost function \(\mathcal{L}\) can often be decomposed into a sum of 
``per-time-step'' cost functions:
\begin{equation}
    \mathcal{L}(\boldsymbol{y}, \boldsymbol{\hat{y}}) = \sum_t \mathcal{L}_t(\boldsymbol{y}_t,\hat{\boldsymbol{y}}_t)
\end{equation}
where, for some activation function \(\boldsymbol{g}\), \(\boldsymbol{y} = \boldsymbol{g}(\boldsymbol{h}_t, \theta)\) are the predicted values and \(\boldsymbol{\hat{y}}\) are the target values.
Computing the gradient with respect to the model parameters \( \theta \), we get:
\begin{equation}\label{eqn:LossGrad}
    \dfrac{\dd \mathcal{L}}{\dd \theta} = \sum_t \dfrac{\dd \mathcal{L}_t}{\dd \boldsymbol{h}_t} \dfrac{\dd \boldsymbol{h}_t}{\dd \theta}.
\end{equation}
The total derivative of the hidden states with respect to the weights ($\dfrac{\dd \boldsymbol{h}_t}{\dd \theta}$) can be computed in a recursive manner:
\begin{equation}
    \dfrac{\dd \boldsymbol{h}_t}{\dd \theta} = \dfrac{\partial \boldsymbol{h}_t}{\partial \boldsymbol{h}_{t-1}} \dfrac{\dd \boldsymbol{h}_{t-1}}{\dd \theta} + \dfrac{\partial \boldsymbol{h}_t}{\partial \theta}.
\end{equation}
We define \( H_t \coloneqq \dfrac{\partial \boldsymbol{h}_t}{\partial \boldsymbol{h}_{t-1}} \in \mathbb{R}^{n \times n}\), \( G_t \coloneqq \dfrac{\dd \boldsymbol{h}_t}{\dd \theta} \in \mathbb{R}^{n \times n \times p}\), \( F_t \coloneqq \dfrac{\partial \boldsymbol{h}_t}{\partial \theta} \in \mathbb{R}^{n\times n\times p} \), and \(G_0\) is initialized with zeros:
\begin{equation}
    \label{eqn:RTRLrecursionrelation}
    G_{t} = H_t G_{t-1} + F_t.
\end{equation}
This is the recursion relation of RTRL \cite{Williams_Zipser89_learalgo}.
Note that the memory requirement of RTRL is independent of the number of time-steps, since it is not necessary to store any intermediate variables.
However, for a simple neuron layer with \(n\) neurons and \(n\) output neurons, the tensor contraction \(H_tG_{t-1}\) has \(\mathcal{O}(n^2p)\) time complexity which becomes \(\mathcal{O}(n^4T)\) for a fully connected layer where \(p\) is the number of parameters per neuron.
Since \(G_t\) has to be stored and updated, memory scales as \(\mathcal{O}(n^3)\).
It is also possible to find the total derivative in an anti-causal manner using a recursion relation that traverses the temporal axis in the opposite direction:
\begin{equation}
    \dfrac{\dd \mathcal{L}}{\dd \boldsymbol{h}_t} = \dfrac{\partial \mathcal{L}}{\partial \boldsymbol{h}_{t+1}} \dfrac{\dd \boldsymbol{h}_{t+1}}{\dd \boldsymbol{h}_{t}} + \dfrac{\partial \mathcal{L}}{\partial \boldsymbol{h}_t}
\end{equation}
Defining \( \boldsymbol{c}_t \coloneqq \dfrac{\dd \mathcal{L}}{\dd \boldsymbol{h}_t} \in \mathbb{R}^n\) and \( \boldsymbol{d}_t \coloneqq \dfrac{\partial \mathcal{L}}{\partial \boldsymbol{h}_t} \in \mathbb{R}^n\), we get:
\begin{equation}
    \boldsymbol{c}_t = \boldsymbol{c}_{t+1} H_{t+1} + \boldsymbol{d}_t
\end{equation}
This is the recursion relation of BPTT \cite{Williams_Zipser89_learalgo}.
Computational complexity is more favorable for BPTT over RTRL because \(\boldsymbol{c}_t\) is a single vector, resulting in a \( \mathcal{O}(n^3T) \) scaling.
It is important to note that we have to compute and store the \( \boldsymbol{c}_t \) at all times \(t\) which leads to a memory scaling linearly with the number of time-steps \(T\), e.g. \(\mathcal{O}(nT)\).
Such a scaling can quickly become problematic for RNNs which require many time-steps, for example in SNN simulations.
Furthermore, the traversal of the computational graph opposite of the temporal axis and the constant accessing of past internal states renders this algorithm none bio-plausible.
RTRL is in principle able to compute gradients \textit{online}, i.e. without requiring an additional backward pass or accessing past internal states, making it closer to how learning might take place in the brain.
However, its substantial computational complexity and $\mathcal{O}(n^3)$ memory, as is shown in \autoref{tab:ComputeMemoryComplexity}, severely limit its practical use.

In this work, we propose an AD framework that leverages the favorable memory scaling and online learning capabilities of RTRL for computations along the time dimension while still using backpropagation to compute the gradient of the network at every step.
We achieve this by taking advantage of the inherent sparsity structure of equation \eqref{eqn:RTRLrecursionrelation} that is well known from other popular sparse approximate gradient-based synaptic plasticity rules \cite{Zenke_Ganguli18_supesupe,Bellec_etal20_soluto,Kaiser_etal20_synaplas}. 
\subsection{Exploiting Sparsity}
\begin{figure}
    \centering
    \includegraphics[width=1\linewidth]{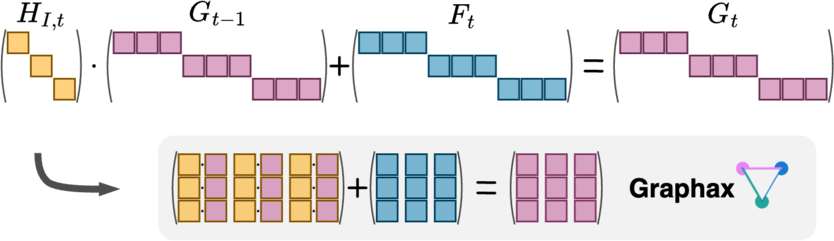}
    \caption{Exploitation of sparsity for the calculation of \(G_{t}\) with Graphax. The sparse matrices are stored as lower dimensional dense representations for less memory and efficient element-wise operations.}
    \label{fig:SparseOps}
\end{figure}

The considerable computational cost of RTRL can be reduced by taking a closer look at the sparse approximations of the tensors \(H_t\) and \(F_t\):
%\(F_t\) is already a diagonal tensor.
We will show that for SNNs \(H_t\) can be decomposed as the sum of a diagonal matrix \(H_{I,t}\) and an off-diagonal matrix corresponding to recurrent connections \(H_{E,t}\).
%% I have doubts about this claim. just remove it for now. 
%Hence, for sparse neural or feed-forward neural networks, both \(H_{I,t}\) and \(F_t\) are sparse as \(H_{E,t}\) gets ignored.
To understand why the sparsity arises, it is helpful to examine what these matrices track.
\(H_{I,t}\) tracks the influence of a neuron on itself and is thereby diagonal by definition. 
Influences by other neurons through recurrent connections are tracked by \( H_{E,t} \).
\(F_t\) measures how the internal state changes with a change in parameters and as such only track the \textit{direct} influence of weights on the the neuron.
Thus, it is also diagonal in networks of fully connected neurons since only synapses of a neuron directly contribute to its state.
In the case of feed-forward networks, \(G_t\) remains diagonal at all times \(t\) as is shown in figure \autoref{fig:SparseOps}. 
We give an example with \(n\) LIF neurons:
\begin{subequations}
    \begin{align}
        \boldsymbol{u}_{t+1} &= \alpha \boldsymbol{u}_t + W \boldsymbol{z}_t \\
        \boldsymbol{z}_{t+1} &= \Theta(\boldsymbol{u}_{t+1} - \boldsymbol{\vartheta})
    \end{align}
    \label{eqn:LIF}
\end{subequations}
where \(\boldsymbol{u}_t\in\mathbb{R}^n\) is the membrane potential at time \(t\), \(\alpha = e^{-\Delta t/\tau_\mathrm{m}}\) is the decay factor for step size \( \Delta t\) and membrane time constant \(\tau_{m}\), \(W \in \mathbb{R}^{m \times n}\) is the synaptic connectivity, \(\boldsymbol{z}_t \in \mathbb{R}^n\) is the spike output, \(\boldsymbol{\vartheta}\in\mathbb{R}^n\) is the spiking threshold, and \(\Theta\) is the threshold function. 
Using the recurrent dynamics on the LIF neuron in equation \eqref{eqn:RTRLrecursionrelation}, we get:
\begin{equation}
    H_t \coloneqq \frac{\dd \boldsymbol{h}_t}{\dd \boldsymbol{h}_{t-1}}
    =\frac{\dd \boldsymbol{u}_t}{\dd \boldsymbol{u}_{t-1}}+\frac{\partial \boldsymbol{u}_t}{\partial \boldsymbol{z}_{t-1}}\frac{\partial \boldsymbol{z}_{t-1}}{\partial \boldsymbol{u}_{t-1}}.
\end{equation}
Thus, we find for the implicit and explicit recurrence that
\begin{equation}
    H_{I,t}\coloneqq\dfrac{\dd\boldsymbol{u}_t}{\dd \boldsymbol{u}_{t-1}} \textrm{ and }
    H_{E,t}\coloneqq\dfrac{\partial \boldsymbol{u}_t}{\partial \boldsymbol{z}_{t-1}}\dfrac{\partial \boldsymbol{z}_{t-1}}{\partial \boldsymbol{u}_{t-1}}.
\end{equation}
\(H_{I,t}\) is a diagonal Jacobian matrix while \(H_{E,t}\) can have any arbitrary sparsity structure.
In practice, \(H_{E,t}\) is often ignored\footnote{this can be for example achieved by wrapping the recurrent connections with a \textit{stop\_gradient} or similar operation or by dropping them in hand-crafted synaptic plasticity rules}, noting that in sparse recurrent networks, sparse k-step-approximations can be used to maintain off-diagonal sparsity in \(G_t\) \cite{Menick_etal20_pracspar}.
%Although we drop some useful information for learning, performance is still comparable to exact gradient methods. 
When $H_{E,t}$ is ignored, \(H_t \coloneqq H_{I,t}\) remains diagonal, inducing local and efficient synaptic plasticity rules \cite{Zenke_Neftci21_brailear}.
However, many typical AD frameworks like PyTorch or JAX are not designed to exploit this type of sparsity in equation \eqref{eqn:RTRLrecursionrelation}. 
In our sparse case, the matrix multiplications in equation \eqref{eqn:RTRLrecursionrelation} can be replaced with a simple element-wise multiplication, reducing the complexity as seen in \autoref{tab:ComputeMemoryComplexity}. 
%The reason being that typical AD frameworks like PyTorch or JAX are not designed to exploit this type of sparsity in Jacobians.
Handcrafted rules could potentially exploit the sparsity, but they require tedious calculations by hand for every new neuron type.
Our work provides a novel type of AD interpreter that exploits the sparsity and resulting favorable compute and memory complexity while being able to automatically handle arbitrary neuron types.

In this work, we will refer to the combination of computing the gradients spatial gradients with AD and accumulating the gradient in time with the approximate RTRL recursion as e-prop.
The reason for this being that both works use the same approximation and we can simply identify the product \( \dfrac{\partial \mathcal{L}}{\partial \boldsymbol{h}_t}G_t \) with an \textit{eligibility trace} \cite{Bellec_etal20_soluto}.
Eligibility traces potentially enable connection between biological learning and spatio-temporal gradient computations common in machine learning.
These traces are comprised of a pre-synaptic and post-synaptic signal as well as a third global factor, as described by three-factor learning \cite{Gerstner_etal18_eligtrac}. 
These traces track the history of neurons' states, thus rendering them local and online. 
%Another proposed approximations changes the way the readout layer is updated: either a local synaptic plasticity rule or random feedback weights to circumvent the weight-transpose problem.
%We do not consider these approximations here since they empirically reduce the performance and the underlying considerations are orthogonal to our investigation.
%
\subsection{Sparse AD with Graphax}
\begin{figure}
    \centering
    \includegraphics[width=0.8\linewidth]{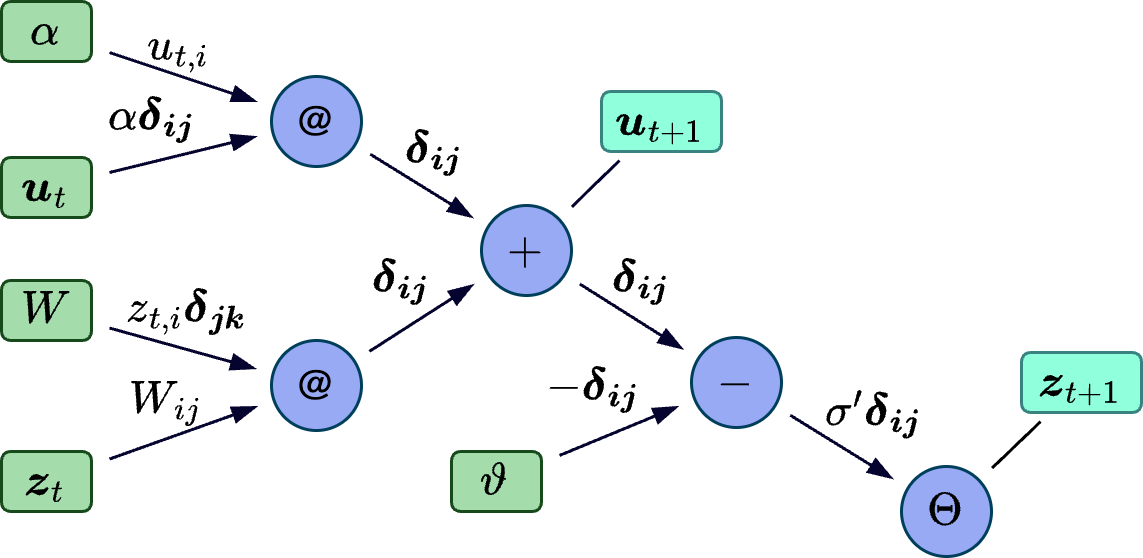}
    \caption{LIF computational graph from equation \eqref{eqn:LIF} with known partial derivatives assigned to each edge as described by \autoref{def:VertexElimination}. Emphasis is made on the Kronecker deltas, which represent diagonal matrices, only edges from \(\alpha\) and \(\boldsymbol{z}_t\) are non diagonal. There are two outputs (in the aquamarine labels), however since \(\boldsymbol{z}_{t+1}\) depends on the \(\boldsymbol{u}_{t+1}\), the computational graph has only one output.}
    \label{fig:CompGraph}
\end{figure}
Graphax is a novel state-of-the-art AD interpreter that can leverage the sparsity of individual Jacobians to accelerate gradient computations \cite{Lohoff_Neftci24_optiauto}. 
It builds on the conceptual framework of tensor-valued vertex elimination that utilizes efficient sparse matrix multiplications and additions. 
\subsubsection{Computing Jacobians with Vertex Elimination}
Vertex elimination is an AD method that accumulates the Jacobian by defining a vertex elimination operation that acts on the computational graph.
We define a computational graph as a tuple \(\mathcal G=(V,E)\) with vertices \(V\) as elemental operations \(\phi_i\) and directed edges \(E\) as outputs \(v_i\) of \(\phi_i\). 
The relation between vertices is stated with \(i\prec j\), where \(i\) has a directed edge connecting it to \(j\) and elemental operation \(\phi_i\) has an output \(v_i\) which becomes the input to \(\phi_j\). 
Every computational graph has as set of input nodes, which only have outbound edges and a set of output nodes, which only have inbound edges. 
We can identify the partial derivative of a function \(\phi_j\) relative to an input \(v_i\), i.e. \(c_{ij}=\frac{\partial\phi_j}{\partial v_i}\), with the corresponding directed edge \((i,j)\in E\).
This enables us to define the vertex elimination operation:
\begin{definition}\label{def:VertexElimination}
\citep{griewank2008evaluating}
For a computational graph \(\mathcal G=(V,E)\) with partial derivatives \(c_{ij}\), \textit{vertex elimination} of \(j\) is defined as
\begin{equation}
    c_{ik}\coloneqq c_{ik}+c_{ij}c_{jk}\quad\quad\forall i, k\in V
\end{equation}
where \(i\prec j\prec k\) and then \(c_{ij}=c_{jk}=0\). Note that by default, any previously undefined partial derivatives \(c_{ik}\) are set to zero.
\end{definition} 
In essence, to eliminate a vertex, we multiply the partial derivative of every inbound edge of the vertex with each partial derivative on the outbound edge.
We create new edges that run from the precursor of the inbound edge to the successor of the outbound edge, thereby bypassing the vertex.
Should the new edge already exist, we just add the values.
The intuitive reason for its mathematical soundness comes from its ``local application of the chain rule''. 
Then we apply vertex elimination consecutively (also called cross-country elimination) on the same graph \(\mathcal G\) until we are left with a bipartite graph. 
The values associated with the edges that connect the input nodes to the output nodes are the components of the resulting Jacobian.
There are two common orders of elimination that Graphax and other AD interpreters implement, forward-mode and reverse-mode. 
Vertex elimination can be used to implement forward-mode AD by eliminating vertices in topologically sorted order.
Reversing this order then gives reverse-mode AD.
\subsubsection{Handling Tensors in Vertex Elimination} 
\begin{figure}
\centering
\begin{subfigure}{0.5\textwidth}
    \centering
    \includegraphics[width=0.6528\linewidth]{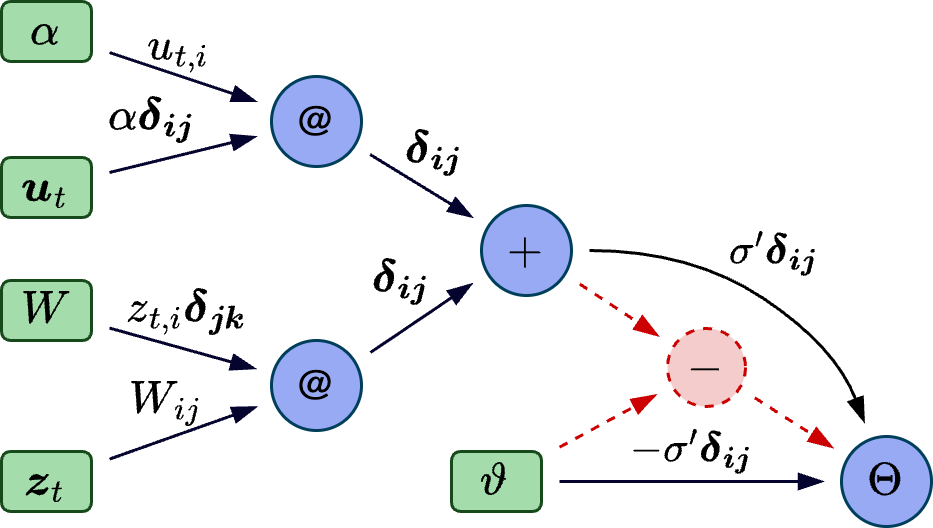}
    \caption{Elimination of the last non-output vertex. We applied the vertex elimination rule as defined below.}
    \label{fig:CompGraphVE1}
\end{subfigure}
\begin{subfigure}{0.5\textwidth}
    \centering
    \includegraphics[width=0.548\linewidth]{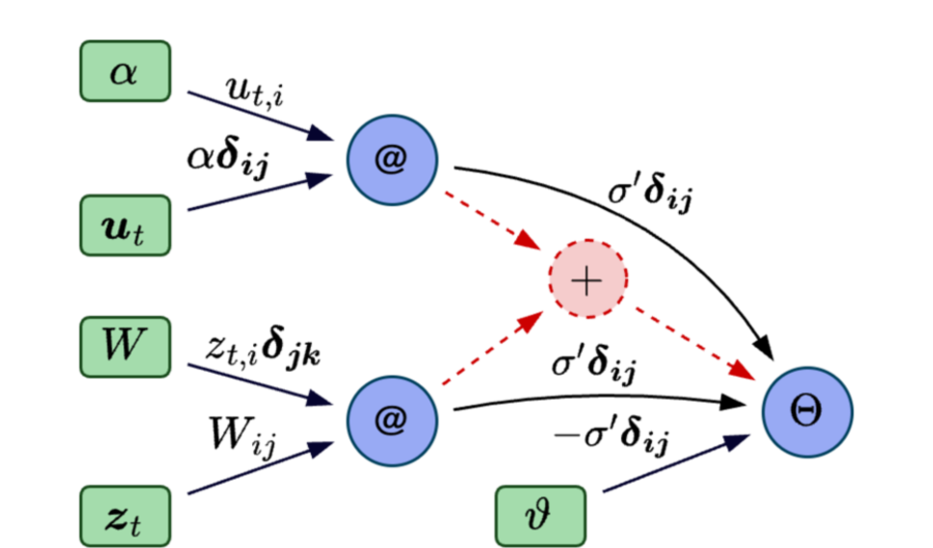}
    \caption{Elimination of the vertex preceeding the last vertex. Proceeding in this order will give use reverse-mode AD. }
    \label{fig:CompGraphVE2}
\end{subfigure}
\caption{LIF computational graph as seen in Figure \autoref{fig:CompGraph} with the first two steps of reverse-mode vertex elimination.\protect\footnotemark The indices used for each Kronecker delta \(\delta_{ij}\) are independent of one another. The independence of the Kronecker delta indices can best be seen between the two steps.}
\label{fig:CompGraphSubfigures}
\end{figure}
\footnotetext{The Heaviside step function \(\Theta\) is not differentiable, therefore we use \(\sigma\) to represent a smooth surrogate function, allowing for differentiation and accounting for represented noise in physical systems \cite{Neftci_etal19_surrgrad}}
In practice, many applications of AD require vector or tensor-valued functions such as element-wise multiplications or matrix multiplications.
In the version presented above, vertex elimination would require a total decomposition of these highly structured operations into their single component operations, which is neither very useful nor computationally efficient.
Instead, we allow the vertices themselves to be vector or tensor-valued, thereby promoting the partial derivatives associated with the edges to Jacobians in their own right which we call partial Jacobians.
\autoref{fig:CompGraph} gives an example of a computational graph for a simple layer of LIF neurons.

Depending on the structure of the operation \(\phi_i\) performed on vertex \(i\), we can infer the sparsity structure of the partial Jacobians.
In most cases, these Jacobians will have a diagonal sparsity structure.
In the LIF neuron layer example in \autoref{fig:CompGraph}, most partial Jacobians are diagonally sparse.
Diagonal sparsity is a form of sparsity in which the non-zero values of a matrix are found exclusively on its main diagonals\footnote{Block-diagonal matrices can also occur in AD, in which partitions of submatrices exist such that only the main-diagonal submatrices are non-zero values. For example this case exists when calculating gradients in a multi-compartment neuron \cite{Zenke_Neftci21_brailear}}. 
For real numbered matrices this is usually written as \(M=\diag(\boldsymbol{v})\) where \(\boldsymbol{v}\in\mathbb{R}^n\) and \(M\in\mathbb{R}^{n\times n}\). As with any kind of sparse matrix, a significant amount of computation can be avoided when multiplying or adding such matrices. For matrix multiplications and additions of size \(n\times n\), computational complexity can go from \(\mathcal{O}(n^3)\) to \(\mathcal{O}(n)\) and from \(\mathcal{O}(n^2)\) to \(\mathcal{O}(n)\) respectively. 
This insight can be extended to tensors of greater dimensions, in which the sparse diagonal can appear in two or more dimensions. 
For example, in a three-dimensional tensor \(T\in\mathbb{R}^{n\times m\times m}\), with the last two dimensions being sparse diagonal, \(T_i=\diag(M_i)\), where \(M\in\mathbb{R}^{n\times m}\) is the non-zero values of \(T\). 
This can be written in simpler terms with the use of the Kronecker delta \(\delta_{ij}\):
\(T_{ijk}=M_{ij}\delta_{jk}\). 
\subsubsection{Efficient Sparse Tensor Contractions}
The \textit{Graphax} package is a JAX-based package that is able perform sparse tensor contractions efficiently and replace expensive generalized matrix multiplications with simple element-wise multiplications if the structure of the partial Jacobians allows.
Graphax stores diagonally sparse tensors in their compressed form, e.g. a diagonal matrix is stored as a vector, a diagonal three-tensor is stored as a matrix etc.
Internally, Graphax keeps track of how the compressed form and the actual tensor shape relate to each other and automatically finds the most efficient way to add or multiply two sparse tensors. 
This leads to significant improvements in runtime and memory utilization.
Since it builds as an additional JAX function transformation, it is fully compatible with all other JAX transformations, including JIT compilation, vectorization, device parallelization and even JAX' own AD library.
We utilize Graphax to implement an efficient version of gradient-based synaptic plasticity rules that are able to exploit the inherent sparsity discussed in previous sections.
We start by fist computing the Jacobians \(H_{I,t}\) and \(F_t\) using the reverse-mode version of vertex elimination so that we get a naturally sparse representation.
\autoref{fig:CompGraphSubfigures} demonstrates how gradients of a simple LIF neuron can be computed using vertex elimination by showing the first two steps of the vertex elimination procedure.
Note that many of the partial Jacobians on the edges have a natural sparsity structure as indicated by the Kronecker deltas \(\delta_{ij}\).
After computing \(H_{I,t}\) and \(F_t\) using the vertex elimination method, we accumulate the gradient forward in time using the recursion relation in equation \eqref{eqn:RTRLrecursionrelation}.
Note that both matrix multiplication and addition take place in the sparse domain using Graphax primitives.
Furthermore, the sparsity shapes, derivatives and gradient accumulation take place completely automatically without additional user input.
Thus we arrive at a truly sparse and general implementation of gradient-based synaptic plasticity.
\begin{table}
\caption{Difference of gradients between BPTT and our work. We report the median and upper 97.5 and lower 2.5 quantiles to quantify the distribution of derivations.}
    \scriptsize
    \begin{center}
        \begin{tabular}{|l|c|c|c|c|}
            \hline
             & Median \( \Delta_\theta \) & Lower \( \Delta_\theta \) (2.5 quantile) & Upper \( \Delta_\theta \) (97.5 quantile)\\
            \hline
            LIF & \( 3.72\cdot 10^{-6} \) & \( 1.01\cdot 10^{-10} \) & \( 5.28\cdot 10^{-5} \)\\
            \hline
            ALIF &\( 4.06\cdot 10^{-6} \) & \( 8.71\cdot 10^{-9} \) & \( 4.95\cdot 10^{-5} \)\\
            \hline
        \end{tabular}
    \label{tab:GradientDeviations}
    \end{center}
\end{table}
\begin{table}
\caption{Test accuracy on SHD of our work compared against BPTT on a feed-forward network with a single hidden layer of spiking neurons with 1024 units and a single non-spiking readout layer to perform the classification task. We only trained the weight matrices of the network and kept the other parameters constant at \(\alpha = 0.95\) and \(\vartheta=1\) for the LIF case. For the ALIF case, we additionally fixed \(\beta = 0.8\).}
    \scriptsize
    \begin{center}
        \begin{tabular}{|l|c|c|}
            \hline
             & LIF & ALIF\\
            \hline
            BPTT & \( 77.3\% \pm 2.3\% \) & \( 83.1\% \pm 1.8 \%\)\\
            \hline
            Ours & \( 78.5\% \pm 2.5\% \) & \( 82.6\% \pm 2.2 \%\)\\
            \hline
        \end{tabular}
    \label{tab:SHDTestAccs}
    \end{center}
\end{table}
\section{Experiments}
In this section, we support our theoretical claims by examining the peak memory usage as well as execution time per time-step.
All of the following experiments were done with a simple feed-forward network with a single hidden layer of LIF neurons with an adaptive threshold (ALIF) and a non-spiking readout layer.
Furthermore, all experiments either used the SHD benchmark data or similar synthetic data with the 700 input channels pooled by a factor of 5 to 140 input channels.  
We start by examining the gradients yielded by the combination of Graphax and the RTRL recursion relation.
In the case of a feed-forward network with a single hidden layer, the gradients of this e-prop-inspired approach should exactly match the gradients computed with BPTT.
The results are demonstrated in \autoref{tab:GradientDeviations} where \( \Delta_\theta \) is the median absolute per-parameter deviation with the upper and lower values being the 2.5\% and 97.5\% quantiles.
The gradients typically agree up to \( 10^{-6} \) since they are both computed up to machine precision.
To further verify the exactness of our method, we trained this simple SNN on the audio benchmark dataset Spiking Heidelberg Digits (SHD) \cite{Cramer_etal20_heidspik}.
\begin{figure}
    \centering
    \includegraphics[width=1\linewidth]{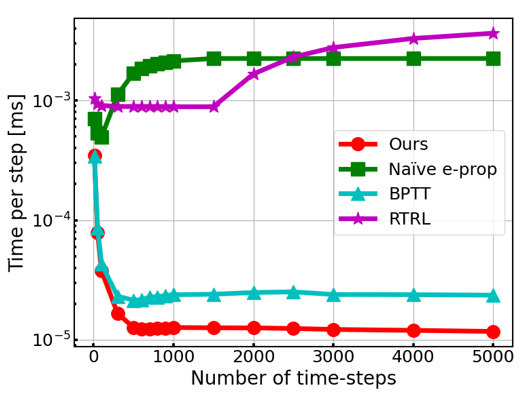}
    \caption{Evaluation time of a single step for 128 hidden neurons with changing training example time-steps for our e-prop implementation with Graphax, BPTT, na\"ive e-prop, and RTRL.}
    \label{fig:TimeTimesteps}
\end{figure}
As would be expected from the equivalence of the gradients, the both training methods achieve the same average test accuracy over 10 training runs with different random seeds.
As discussed in \autoref{sec:Eprop}, our work is expected to have a more favorable performance when compared with respect to number of time-steps compared to BPTT and with respect to number of neurons compared to RTRL. 
In this work we verify these expectations and also demonstrate that leveraging the inherent sparsity of \(H_{I,t}\), \(G_{t}\) and \(F_{t}\), as given in equation \eqref{eqn:RTRLrecursionrelation}.
In the following experiments, we typically compare four different cases to demonstrate the efficacy of our approach:
\begin{itemize}
    \item \textbf{BPTT} 
    We train the the SNN by differentiating the \textit{for}-loop with reverse-mode AD using the JAX \textit{jax.lax.scan} and \textit{jax.jacrev} primitives. Since we only have a feed-forward network with a single hidden layer of spiking neurons, BPTT and e-prop gradients agree.
    \item \textbf{RTRL} 
    The setup is the same as in BPTT, but instead of reverse-mode AD, we use forward-mode with \textit{jax.jacfwd}.
    \item \textbf{Na\"ive e-prop} 
    We implement the recursion relation explicitly into the \textit{for}-loop and compute the gradients of \(H_{I,t}\) and \(F_{t}\) using \textit{jax.jacrev}. 
    The experiment is necessary to show that the exploitation of the inherent sparsity in the recursion relation leads to a signification increase in performance. 
    Note that in this case, we do the full tensor-contraction of equation \autoref{eqn:RTRLrecursionrelation}.
    \item \textbf{Our work} 
    Our work uses Graphax' sparse AD primitive \textit{graphax.jacve} to directly compute sparse gradients using vertex elimination. 
    We then implement the recursion relation using these sparse gradients to accelerate the gradient computations and reduce memory consumption.
\end{itemize}
We set out to verify both the favorable compute and memory scaling by performing two experiments for each case.
The first aimed to test different sequence lengths, where we fixed the number of hidden neurons to 128 and varied the number of time-steps from 10 to 5,000 in increments of 10, 100, 500, and 1000.
The second aimed to test different number of hidden neurons, where we fixed the time-steps to 1,000 and varied the number of hidden neurons from 16 to 512 in increments of powers of two.
\begin{figure}
    \centering
    \includegraphics[width=1\linewidth]{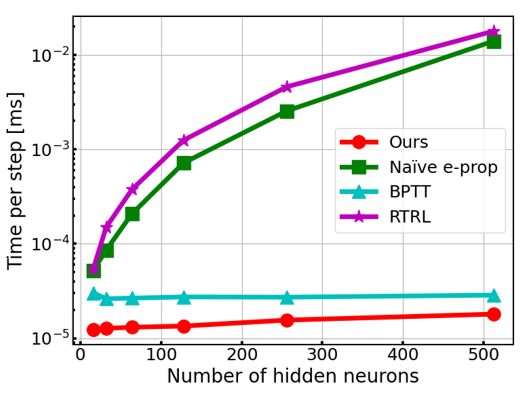}
    \caption{Evaluation time of a single step for changing number of hidden neurons with 1000 training example time-steps for our e-prop implementation with Graphax, BPTT, na\"ive e-prop, and RTRL.}
    \label{fig:TimeHidden}
\end{figure}
\subsection{Execution Time}
We measured computational complexity as the average number of milliseconds for a single time-step.
First we measured computational complexity for different sequence lengths as seen in \autoref{fig:TimeTimesteps}. We used a logarithmic scale for the time axis. We expected that they are constant for different sequence lengths.
However, we observed a significant change from 2 time-steps to approximately 500 time-steps across all implementations. 
Since we used loop-unrolling in the \textit{jax.lax.scan} primitive which enables the XLA compiler used by JAX to create a single kernel for multiple time-steps at once, we presume that this unrolling together with the scheduling of such operations on the GPU might be causing the observed behavior.
For longer sequences, the time per step becomes approximately constant for three of four cases, with the exception of RTRL.
This counter-intuitive behavior is intriguing, but a detailed analysis, possible at the compiler level, is out of the scope of this publication.
We notice that the na\"ive implementation performs over two orders of magnitude worse than our implementation, strongly supporting our claim that the exploitation of sparsity can lead to significant gains in runtime.
Surprisingly, our implementation even seems to outperform the commonly used BPTT. 

Secondly we measured computational complexity for different number of hidden neurons as seen in \autoref{fig:TimeHidden}. 
We again used a logarithmic scale for the time axis.
We see our implementation outperforms all others again, including BPTT up to 512 hidden neurons.
The time per step for BPTT appears nearly constant, which may be attributed to parallelization of computations across neurons. 
%However, this behavior needs further investigation to confirm our explanation. 
RTRL and na\"ive e-prop display an exponential increase in time per step as neuron count is increased, which is unfavorable compared to BPTT and our implementation.
\begin{figure}
    \centering
    \includegraphics[width=1\linewidth]{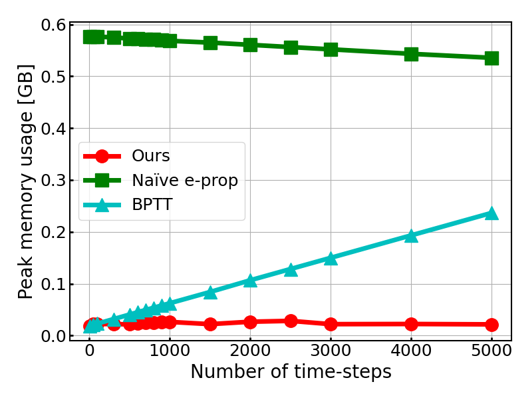}
    \caption{Peak device memory usage for 128 hidden neurons with changing training example time-steps for our e-prop implementation with Graphax, BPTT, and na\"ive e-prop.}
    \label{fig:MemoryTimesteps}
\end{figure}
\subsection{Peak Memory Usage}
The first experiment shows peak memory usage with a fixed number of hidden neurons (\autoref{fig:MemoryTimesteps}). 
The peak memory usage for RTRL was at around 3.6 GB and stayed constant across time-steps (not shown in figure). 
As expected, BPTT depends linearly on the number of time-steps, while our implementation, like RTRL, is constant in memory. 
An unexpected result came from the na\"ive implementation, in which it slowly decreased in peak memory usage. 
We speculate that this is due to loop optimizations in the XLA compiler.
Nevertheless, our implementation outperforms all of the others, beating BPTT across the board due to its constant memory requirements.

\begin{figure}
    \centering
    \includegraphics[width=1\linewidth]{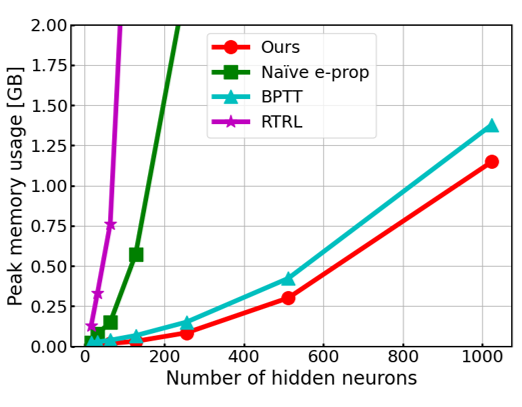}
    \caption{Peak device memory usage for changing number of hidden neurons with 1,000 training example time-steps for our e-prop implementation with Graphax, BPTT, and na\"ive e-prop.}
    \label{fig:MemoryHidden}
\end{figure}

\autoref{fig:MemoryHidden} shows peak memory usage as a function of the number of hidden neurons with fixed number of time-steps. 
We expect the peak memory usage of BPTT and our implementation to scale approximately linearly with the number of hidden neurons. 
Meanwhile, the na\"ive implementation and RTRL are expected to have a quadratic increase with their peak memory usage with increasing number of neurons as seen in \autoref{tab:ComputeMemoryComplexity}. 
We observe a slightly superlinear scaling among the results for BPTT and our implementation, but still verify that both significantly outperform RTRL and the na\"ive e-prop implementations. 
Although the peak memory usage between our implementation and BPTT is at most 250 MB, our implementation remained consistently lower across different numbers of hidden neurons.
\section{Conclusion}
We demonstrated a general, sparsity-aware gradient-based synaptic plasticity, which shows improved compute and memory complexity across scales.
Namely, we demonstrated the superior \textit{constant-in-time} memory scaling of our approach similarly to RTRL, which enables the training spiking neuron models on long sequences.
This is in stark contrast to many of the current state-of-the-art implementations that rely on BPTT and do not utilize the inherent sparsity, whose memory grows with sequence length 
%Therefore, with memory proportionally increasing, it often suffers from a significant memory overhead.
Furthermore, our Graphax AD framework made our implementation scalable and generalizable across different hardware backends and arbitrary neuron types (e.g. LIF and ALIF) without putting any additional requirements on the user.
Thus, our work enables bio-plausible synaptic plasticity at scale with AD that truly exploits the features of the underlying approximations.
While all our experiments were performed on a simple feed-forward network with a single hidden layer of spiking neurons and a non-spiking readout layer, they can be straight-forwardly generalized to the multi-layer case.
In this case, the equivalence of the gradients with BPTT is no longer true. 
However, a rich body of past work has addressed this issue already and successfully illustrated that this does not significantly inhibit learning\cite{Kaiser_etal20_synaplas, Bohnstingl_etal20_onlispat}.
Generalizing our approach to the multi-layer case is an exciting research avenue that we intent to pursue in future work.
\newpage
\bibliographystyle{plainnat}
\bibliography{refs,biblio_unique_alt}

\end{document}